\documentclass[twoside,leqno,twocolumn]{article}  
\usepackage{ltexpprt}

\usepackage{amsmath,amssymb, graphicx}
\usepackage[ruled,linesnumbered]{algorithm2e}
\usepackage{algorithmic}
\usepackage{paralist} 
\usepackage{color}
\usepackage{multirow}
\usepackage{rotating}
\usepackage{hyperref}
\usepackage[mathscr]{eucal} 
\usepackage{amsbsy} 
\usepackage{booktabs}
\usepackage{xcolor}
\usepackage{tikz}
\usetikzlibrary{snakes}
\usepackage{multirow}
\usepackage{epstopdf}
\usepackage{balance}
\usepackage{tablefootnote}
\usepackage{empheq} 
\usepackage{moresize}
\usepackage{enumitem}
\setlist{nolistsep}
\usepackage{graphicx}
\usepackage{subcaption}
\usepackage{multicol, blindtext}
\usepackage{amssymb}
\usepackage{pifont}
\newcommand{\cmark}{\ding{51}}%
\newcommand{\xmark}{\ding{55}}%

\usepackage[font=small,labelfont=bf,skip=0pt]{caption}

\DeclareCaptionType{copyrightbox}
\usepackage{url}

\newcounter{ALC@tempcntr}
\newcommand{\hide}[1]{}

\newcommand{\ben}{\begin{enumerate*}}
\newcommand{\een}{\end{enumerate*}}
\newcommand{\bit}{\begin{itemize*}}
\newcommand{\eit}{\end{itemize*}}

\begin{document}

\author{
Uday Singh Saini\\
       University of California Riverside\\
       {usain001@ucr.edu}
\and
Evangelos E. Papalexakis\\
       University of California Riverside\\
       {epapalex@cs.ucr.edu}
}
\date{}
\title{Analyzing Representations inside Convolutional Neural Networks}

\maketitle
\begin{abstract}

How can we discover and succinctly summarize the concepts that a neural network has learned? Such a task is of great importance in applications of networks in areas of inference that involve classification, like medical diagnosis based on fMRI/x-ray etc. In this work, we propose a framework to categorize the concepts a network learns based on the way it clusters a set of input examples, clusters neurons based on the examples they activate for, and input features all in the same latent space. This framework is unsupervised and can work without any labels for input features, it only needs access to internal activations of the network for each input example, thereby making it widely applicable. We extensively evaluate the proposed method and demonstrate that it produces human-understandable and coherent concepts that a ResNet-18 has learned on the CIFAR-100 dataset. 
\end{abstract}
\section{Introduction}
\label{sec:intro}

With the advent of deep neural network architectures as the prominent machine learning paradigm \cite{NIPS2012_4824} for human-centric applications a common issue that has plagued their application is the lack of interpretability of these models. As the spectrum of domains where deep learning replaces traditional and orthodox methods expands, and deep learning methods percolate to areas of immediate applicability to daily life, like self driving cars\cite{DBLP:journals/corr/BojarskiTDFFGJM16}, understanding what networks do takes on a more central role than aspiring performance gains. Future challenges that machine learning engineers face, are not just limited to improving model accuracy, but also debugging\cite{Tan_2018} and training networks in order to make them conform to ever evolving regulations concerning ethics\cite{IR} and privacy\cite{doi:10.1098/rsta.2016.0118}.

Most literature in the area of explainable AI focuses on providing explanations for pre-trained networks\cite{lime},\cite{ghorbani2019automatic}. While some methods focus on designed models which have explainability as a part of their design philosophy\cite{alvarezmelis2018robust}. Our work belongs to the former category and focuses on providing explanation for already trained models, or what is colloquially called post-hoc explanation. Within the strata of post-hoc explanations, there exist multiple evolutionary branches, some focus on interpreting the features\cite{SanityNIPS2018}, and\cite{Zhou2018InterpretableBD} interprets the network by breaking down an input prediction into semantically interpretable components and works like \cite{8417924} focus on interpreting neurons based on their behaviour when they activate for entities like different textures, colours and images.\\
We focus on unsupervised discovery of concepts learned by the network by trying to cluster the neurons, input features and inputs themselves in the same latent space. The motivation for doing so comes from works like\cite{DBLP:journals/corr/GardnerKLUWH15} where it has been conjectured that natural images usually lie on a manifold and that a neural network embeds this manifold as a subspace in it's feature space. The work most similar in spirit to ours is ACE\cite{ghorbani2019automatic} where the goal is to explain the prediction of neural networks not in terms of individual neurons, but rather, by focusing on learning the concepts utilized by the network that are most sensitive for a successful prediction, and learning of such concepts is a supervised process. Unlike our work, ACE\cite{ghorbani2019automatic} utilizes existing algorithms or manual annotation to curate a set of concepts, feed it to the network and measure the sensitivity of the network to those concepts using TCAV\cite{kim2017interpretability}. This solution, though elegant relies heavily on domain expert annotators or supervised tools, while we learn these concepts from the activations of the network and try to determine concepts learned by the network by probing it input examples. Another line of work in \cite{kim2017interpretability} focuses on learning vectors which when measured for their effects on class prediction, align with high-sensitivity directions in the latent space of the network. We also utilize \cite{kim2017interpretability} as a means to validate our approach in \autoref{sec:ext}.


Our approach aims to find a latent representation for neurons, input features and examples in a common subspace, where clustering them aims to elicit meaningful insights about the networks ability to discern between examples. Using such a tri-factor clustering, we can analyze intersections between groups of neurons which fire for different classes, focus on which input features provide a basic structure upon which the model correctly classify its inputs and  analyze an individual example based on their similarity and differences to other examples, as determined by the network's embeddings of them. We model our problem as a coupled matrix factorization, where the model is constrained to appropriate constraints like non negativity, which aid in interpretability \cite{5c8e335ac9274d0aad4b96178bf24394} and the possibility of adding regularizations like group sparsity, orthogonality etc. to encode meaningful priors into the model. We conduct our analysis by observing the behaviour of a network on a set of images it has previously not seen, for the purpose of this study, we experiment with CIFAR-10 and CIFAR-100 as our datasets of choice.  Our raison d'etre is to approach the problem of concept discovery in an unsupervised manner, in order to bridge a gap unfulfilled by \cite{ghorbani2019automatic} and \cite{8417924}. In doing so develop a methodology which can seed or supplement other interpretability methods.

\section{Related Work}
\label{sec:RelatedWork}
In our work we aim to interpret a learned model using a set of images which may or may not have been a part of the set of training classes of the network. Our work comes in stark contrast with most existing literature, since the goal in this work is not to evaluate the network on a feature by feature or on a sample by sample basis as in \cite{zeiler2013visualizing},\cite{koh2017understanding}, \cite{smilkov2017smoothgrad},\cite{simonyan2013deep}. Additionally, there are other works, such as \cite{zeiler2013visualizing} which visualize a network based on images that maximize the activation of hidden units or works like \cite{mahendran2014understanding} which use back-propagation to generate salient features of an image. \\
Works like \cite{alvarezmelis2018robust} focus on explaining a network by proposing a new framework where the network is forced to learn concepts and demonstrate their relevance towards a prediction. This framework relies on prior constraining and encoding for what is thought to be a concept. In \cite{Zhou2018InterpretableBD} the focus is on explaining each prediction made by the network by decomposing the activations of a layer in the network into a basis of pre-defined concepts, where each explanation a weighted sum of these concepts, where the weights determine the impact each concept has towards prediction.
Our work has similarities of philosophy with the previous 2 works, but unlike \cite{alvarezmelis2018robust} we don't focus on learning an interpretable model, instead we focus on unsupervised explanation of an already trained network. And unlike \cite{8417924} we do not have a pre-made notion of concepts, instead we let the model learn underlying concepts based on the set of examples fed in the analysis. This way our approach is application agnostic.
Recent work on Network Dissection \cite{DBLP:journals/corr/BauZKOT17} tries to provide a framework where they can tie up a neuron in the network to a particular concept for which the neuron activates. These concepts can be simple elements like colour, to compound entities like texture. They accomplish this through a range of curated and labeled semantic concepts whereas our work doesn't need user labeled data. 
Another work which relies on interpreting the network through the lens of abstract concepts is TCAV \cite{kim2018interpretability}. This work tries to provide an interpretation into network's workings in terms of human interpretable concepts. Like our work, they too rely on the internal representation of the network to determine the network's behaviour, but unlike us they utilize manual/pre-defined concepts and test the network's sensitivity towards it.
The work presented in \cite{raghu2017svcca} uses a variant of canonical co-relational analysis and focuses on learning the complexity of the representations learned by the network to determine the dynamics of learning, our work differs as we use the structure of the learned representation as a guideline for our factorization framework and don't comment on the inherent complexity.\\ 
The work most in line with our goals is \cite{ghorbani2019automatic}, here the authors seek to automatically discover concepts learned by the network which are of high predictive value, as measured by their TCAV score \cite{kim2018interpretability}.\\
In \autoref{fig:comp} we compare our work to other works in the area, some of which relevant and others more tangential to our approach. While the axioms of interpretable machine learning are an ever evolving set of principles, we enlist a few features that help us highlight the differences between our work and it's closest neighbours in this space. Our work is the only unsupervised method in this space of model interpretability which helps us discover concepts learned by the network in terms of the examples clustered by the network. ACE \cite{ghorbani2019automatic}, SeNN\cite{alvarezmelis2018robust} learn concepts but either by utilizing explicit supervision or by employing pre-existing trained models, whereas works like \cite{8417924} require detailed human labeling of neurons and image pixels and patches, thus making the process slow and sluggish for adaptation to a new domain. LIME \cite{lime} on the other hand tries to visualize a linear decision boundary across an input, which we approximate by the input's K-Nearest Neighbours but unlike our work it cannot discover abstract concepts learned by the network without significant modifications.

\begin{figure*}
	\caption{Relevant Work Comparison}
	\label{fig:comp} 
	\centering
	\begin{tabular}{|c|c|c|c|c|c|} 
		
		\hline
		Model Features & ACE\cite{ghorbani2019automatic} & LIME\cite{lime} & SeNN\cite{alvarezmelis2018robust} & Net Dissection \cite{8417924} & Our work  \\
		\hline
		Post-Hoc Interpretability & \cmark & \cmark & \xmark & \cmark &\cmark \\
		\hline
		Unsupervised & \xmark & \xmark & Partial & \xmark & \cmark \\
		\hline
		Insights on Inputs & \cmark & \cmark & \cmark & \cmark & \cmark \\
		\hline
		Insights on Neurons & \xmark & \xmark &\xmark & \cmark & \cmark \\
		\hline
		Insights on Features & \xmark & \cmark & \cmark & \cmark & \cmark \\
		\hline
		Collective Analysis of Inputs & \xmark & \xmark & \xmark & \xmark & \cmark \\
		\hline
		Individual Analysis of Inputs & \cmark & \cmark & \cmark & \cmark & Partial \\
		\hline
		Analysis of Representation Space & \xmark & \xmark & \xmark & \xmark & \cmark \\
		\hline
	\end{tabular}
\end{figure*}

\section{Proposed Method}
\label{sec:method}
In this section we begin by outlining the motivation for our methodology, we then proceed to outline the implementation schema and optimization problem for our model. Subsequently we present the model details and lay down the groundwork for evaluation protocols suited for this method.


\subsection{Motivation}


Our goal is to visualize the latent representation space learned by a Neural Network by comparing and contrasting the behaviour of the network on different types of inputs. We want to accomplish this in a framework where we can explain the concepts learned by the network in terms of the inputs that are used to probe the network. In doing so we can assess the generalization ability of the network, both to familiar and unseen datasets, thus providing insights to human evaluators about the health of the trained network and it's suitability to a particular domain. This is possible because there are no restrictions on what qualifies as a legitimate dataset for evaluating network behaviour, thus in theory, we can evaluate a network on a dataset which is different from it's training dataset and assess the suitability of the architecture to learn atomic concepts (which may be valid across domains) from the training data instead of learning it's idiosyncrasies.


\subsection{Proposed Model}
Given these goals in mind, we lay down the model principles aligned with our objectives. Our approach is a method that relies on a coupled matrix factorization framework where we compute embeddings of test examples and individual neurons in probed layers in a shared latent space. Our method relies on only having access to activations of internal layers of a network for a given input. Additionally, for ease of modelling, we assume that these activations are non-negative in nature, for instance  ReLU and Sigmoid non-linearities are used in the network. In doing so our model does not introduce any external learning constraints while training the network, thus lending it universality. We probe various layers of a network with a set of test examples, and for each test example, we store the network's response across all (observed) layers.  We do so with an aim to breakdown the process of interpretability into a process of finding common local structures across various test/evaluation examples, where each feature in the latent representation hopefully captures a latent semantic concept. Thus, through the lens of our model, we can, hopefully view individual concepts in an amalgam of constituents of a test example. In the following subsections we describe model construction and provide mathematical details of implementation.


\subsubsection{Model Construction}
For our analysis we need construct a set of matrices where each matrix $A_i$ in the set is a matrix $\in$ $ \mathbb{R}^{a_i  \times  N}_{+}$, where $a_i$ is the number of neurons in layer $i$ of the network, and $N$ is the number of examples on which our analysis is conducted. Each column $k$ of matrix $A_i$, is a vectorized activation of layer $i$ of the network for a given test sample $k$. Thus, to reiterate, a column $k$ of this matrix $A_i$, denoted by $A_i[:,k]$, is the activation of layer $i$ of the network when the $k^{th}$ test example is passed as an input to it.
Along similar lines we construct another set of matrices where each Matrix $D_i$ $\in$ $ \mathbb{R}^{S_i  \times  N}_{+}$ where $S_i$ is number of pixels in the $i^{th}$ channel of input images and $N$ is the same as earlier. On the same lines as before, each column $k$ of matrix $D_i$, is the $k^{th}$ test sample's $i^{th}$ channel vectorized.

\subsubsection{Model}
The objective function for our proposed method is as follows:
 \begin{equation}\begin{split}J(\textbf{P},F,\textbf{O}) = \sum_{i = 0}^{C-1} \lVert D_i - P_iF\rVert_{F}^{2}   +  \sum_{j = 0}^{L-1} \lVert A_j - O_jF\rVert_{F}^{2} +\\ \sum_{i = 0}^{C-1}\lambda_P\lVert P_i \rVert^{2}_{p} + \sum_{j = 0}^{L-1}\lambda_O\lVert O_j \rVert^{2}_{p} + \lambda_F\lVert F \rVert^{2}_{p} \\ \ni P_{i}, O_{j}, F, \in \mathbb{R}^{S_i \times d}_{+},\mathbb{R}^{N_j \times d}_{+},\mathbb{R}^{d \times N}_{+}\\ ||P[:,i]||^{2}_{2} = 1 , ||O[:,i]||^{2}_{2} = 1 \forall  i \end{split}\label{eq:1}\end{equation}
 In \autoref{eq:1}, C is the number of channels in input data, L is the number of layers of the network that are part of analysis -  as we can select the non-negative layers we want to analyze and are not obligated to include all the layers of any architecture. $p$ is usually 2 for $2-Norm$ regularization although for the purposes of some experiments we instead set the column norms of the Pixel and Neural Factor matrices to 1 \\
 For each matrix $D_i$ in \autoref{eq:1}, it's $k^{th}$ column is the input data's channel $i$ vectorized as input. Thus for instance, for a 3-channel image, with image number $j$ of the test set, $D_0[:,j]$ is the vectorized $0^{th}$ channel of the $j^{th}$ image and so on. \\
 As mentioned earlier, each $D_i$ is thus a matrix of Pixel-by-Example. Each $P_i$ in the first term of the summation in equation \ref{eq:1} is a latent representation matrix for each pixel. That is, Each row of $P_i$, for instance $P_i[k,:]$ is the latent representation of the $k^{th}$ pixel in the input space. \\
 For each matrix $A_j$ in \autoref{eq:1}, it's $k^{th}$ column is the activation of layer $j$ of the network for $k^{th}$ test input. Thus for instance, image number $j$ of the test set, $A_0[:,j]$ is the activation of layer $0$ of the network for image $j$, $A_1[:,j]$ is the activation of layer $1$ of the network for image $j$, $A_2[:,j]$ is the activation of layer $2$ of the network for image $j$, as a point of caution we would like to mention that  $A_0$,$A_1$,$A_2$ and so on, do not necessarily correspond to layers 0,1,2 of the network, they instead correspond to the $0^{th}$,$1^{st}$,$2^{nd}$ analyzed layers of the network, as our model offers the ability to skip layers of the network, in our convention, the higher index of the layer, the deeper we are in the network.\\
Each matrix $A_j$ encodes the activity of neurons of layer $j$ for a given training example. Therefore, each $O_j$ in the factorization encodes the latent representation of neurons of layer $j$ in it's rows. That is, $O_j[k,;]$ is the latent representation of $k^{th}$ neuron of layer $j$.
Similarly, the matrix $F$ encodes in it's columns, the latent representation of each test example fed to the network. That is, $F[:,k]$ is a d-dimensional latent representation of test sample $k$.
Each factor matrix in the objective function obeys non-negativity constraints, and we use multiplicative update rules as described in \cite{Lee:2000:ANM:3008751.3008829} to solve for the factor matrices. 

 Update Steps for solving the factor matrices in Equation \ref{eq:1} are as presented in the following Equation \autoref{fig:UpdateSteps} :-
 \begin{eqnarray*}\label{fig:UpdateSteps}
  F &\leftarrow& F * \frac{\sum\limits_i P_{i}^TD_{i} +\sum\limits_j O_{j}^TA_{j}}{\sum\limits_i P_{i}^TP_{i}F +\sum\limits_j O_{j}^TO_{j}F + \lambda_{F}F} \\
  P_i &\leftarrow& P_i * \frac{D_{i}F^T}{P_iFF^T+ \lambda_{P}P_i} \\
 O_j &\leftarrow& O_j * \frac{A_jF^T}{O_jFF^T + \lambda_{O}O_j}
\end{eqnarray*}

\subsubsection{Model Intuition}
We now provide some intuition for our modeling choices. Our goal is to identify hidden patterns or concepts that the network learns in order to classify data. To achieve this our model clusters the test examples, neurons and pixels in the same inner product space. We achieve this clustering by incorporating a coupled non-negative matrix factorization framework. 
In our learned vector representation of these 3 types of objects, a high value along a latent dimension indicates that a particular latent concept participates in explaining the behaviour of the object. By constraining the model to adhere to a non negative framework, we encourage an interpretable sum-of-concepts based representation\cite{5c8e335ac9274d0aad4b96178bf24394}.

Further elaborating on the learned factor matrices, Each column $j$ of Matrix $P_i$ $\in$ $ \mathbb{R}^{S_i  \times  d}_{+}$ is the activation of the pixels of channel $i$ for the concept discovered in latent factor $j$. Collecting such information over all input channels $i$ for a given $j$ in the respective factor matrices we can uncover the average activation of pixels across channels for a given concept. This representation can be thought of as a channel-wise mask over features in the input, similar to LIME \cite{lime}, but instead we discover a latent concept level mask as opposed to an input level mask.
Matrix $F$ $\in$ $ \mathbb{R}^{d  \times  N}_{+}$ is the input representation matrix where each column $k$ of $F$ is a vector in $\mathbb{R}^{d}_{+}$ representing the $k^{th}$ example in the same latent space as Pixels and Neurons. For any input $k$, A high value along any component $j$ of it's d-dimensional representation indicates a high affinity of this input towards the latent concept encoded in the dimension $j$ and $P_0[:j]$,$P_1[:j]$,$P_2[:j]$ together will us visualize the pixel activation mask for this latent concept $j$ as discussed earlier. Collecting all the highest affinity inputs for each latent factor, we obtain a visual approximation of the concept learned in this latent dimension. Given the unsupervised nature of this model, it extremely well suited for concept discovery for neural networks, akin to a similar role played by  ACE \cite{ghorbani2019automatic} for TCAV \cite{kim2017interpretability}.
Matrices $O_j$'s embed neurons of a layer $j$ in the same latent space as inputs and features and help us visualize which neurons in a layer activate for which concept, we do this by demonstrating the similarity of latent concepts when measured w.r.t. neurons of a layer. We can also look at the behaviour of neurons across layers by observing the cohesiveness of latent space as the neurons go deeper in the network.  

\section{Experimental Evaluation}
\label{sec:experiments}
In the following  subsections we will present the analysis of the latent space learned by a ResNet-18 \cite{DBLP:journals/corr/HeZRS15} when trained on CIFAR-100 images \cite{Krizhevsky09learningmultiple}. Our analysis touches all the modalities captured by our model, i.e. Analysis of Pixels, Analysis of Neurons and Analysis of Examples. We present this analysis in 3 subsections for a given network. We also released the code\footnote{Code: https://github.com/23Uday/Project1CodeSDM2021} for verification.
\subsection{Analysis of A ResNet-18 on CIFAR-100 Dataset:}
In the following subsections we analyze the behaviour of a ResNet-18\footnote{https://github.com/kuangliu/pytorch-cifar} trained and analyzed on CIFAR-100. Each subsection represents a modality of analysis, namely, inputs, Neurons, and input features or pixels themselves.
\subsubsection{Analysis of Input representations:}
\label{sec:concepts}
In this section we present the analysis of representations learned in the input representation Matrix $F$. For each latent dimension $i$ we compute the total class-wise activation score of inputs in the row $F[i,:]$ and present the top 3-4 activated classes along that latent dimension, the images which had the highest affinity in this latent dimension and most activated super-class in Table \ref{tab:Table1}. The motivation behind analyzing super class labels is to validate the assertion that each latent factor captures an abstract concept that is predominantly present in the member images. We reiterate that these super class labels were not used in training of the network but only used as a means to assign a pseudonym to each latent factor, the validity of which can be verified by looking at the topmost activated images and the group of top most activate classes in Table \ref{tab:Table1}.
\begin{table*}[!ht]
	\caption{Matrix-$F$ Latent Factor Analysis For ResNet-18} \label{tab:Table1}
	\begin{tabular}
		{l|c|c} \hline Factor: Top Classes  & Top Images & Top 1-2 Super Class\\
		\hline  0: bed,television,wardrobe  & \includegraphics[width=0.4cm]{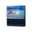}\includegraphics[width=0.4cm]{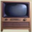}\includegraphics[width=0.4cm]{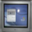}\includegraphics[width=0.4cm]{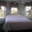}\includegraphics[width=0.4cm]{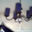}\includegraphics[width=0.4cm]{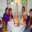}\includegraphics[width=0.4cm]{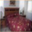}\includegraphics[width=0.4cm]{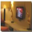}\includegraphics[width=0.4cm]{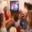}\includegraphics[width=0.4cm]{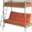}\includegraphics[width=0.4cm]{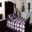}\includegraphics[width=0.4cm]{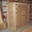}\includegraphics[width=0.4cm]{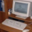}\includegraphics[width=0.4cm]{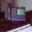}\includegraphics[width=0.4cm]{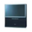}\includegraphics[width=0.4cm]{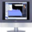}\includegraphics[width=0.4cm]{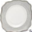} & household furniture\\
		\hline  1: kangaroo,beaver,bear  & \includegraphics[width=0.4cm]{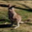}\includegraphics[width=0.4cm]{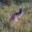}\includegraphics[width=0.4cm]{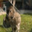}\includegraphics[width=0.4cm]{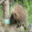}\includegraphics[width=0.4cm]{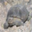}\includegraphics[width=0.4cm]{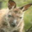}\includegraphics[width=0.4cm]{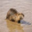}\includegraphics[width=0.4cm]{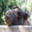}\includegraphics[width=0.4cm]{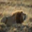}\includegraphics[width=0.4cm]{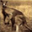}\includegraphics[width=0.4cm]{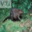}\includegraphics[width=0.4cm]{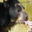}\includegraphics[width=0.4cm]{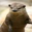}\includegraphics[width=0.4cm]{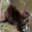}\includegraphics[width=0.4cm]{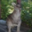}\includegraphics[width=0.4cm]{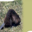}\includegraphics[width=0.4cm]{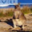}&large omnivores and herbivores\\
		\hline  2: mountain,castle,bridge  & \includegraphics[width=0.4cm]{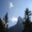}\includegraphics[width=0.4cm]{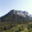}\includegraphics[width=0.4cm]{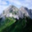}\includegraphics[width=0.4cm]{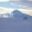}\includegraphics[width=0.4cm]{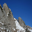}\includegraphics[width=0.4cm]{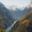}\includegraphics[width=0.4cm]{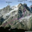}\includegraphics[width=0.4cm]{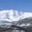}\includegraphics[width=0.4cm]{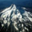}\includegraphics[width=0.4cm]{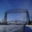}\includegraphics[width=0.4cm]{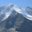}\includegraphics[width=0.4cm]{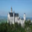}\includegraphics[width=0.4cm]{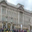}\includegraphics[width=0.4cm]{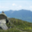}\includegraphics[width=0.4cm]{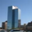}\includegraphics[width=0.4cm]{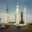}\includegraphics[width=0.4cm]{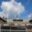}&large man made outdoor things\\
		\hline  3: willow, maple, pine, oak  & \includegraphics[width=0.4cm]{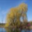}\includegraphics[width=0.4cm]{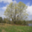}\includegraphics[width=0.4cm]{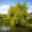}\includegraphics[width=0.4cm]{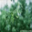}\includegraphics[width=0.4cm]{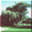}\includegraphics[width=0.4cm]{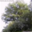}\includegraphics[width=0.4cm]{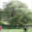}\includegraphics[width=0.4cm]{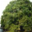}\includegraphics[width=0.4cm]{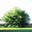}\includegraphics[width=0.4cm]{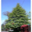}\includegraphics[width=0.4cm]{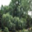}\includegraphics[width=0.4cm]{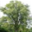}\includegraphics[width=0.4cm]{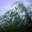}\includegraphics[width=0.4cm]{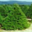}
		\includegraphics[width=0.4cm]{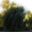}\includegraphics[width=0.4cm]{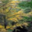}\includegraphics[width=0.4cm]{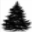}&trees\\
		\hline  4: shark,dolphin,whale  & \includegraphics[width=0.4cm]{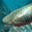}\includegraphics[width=0.4cm]{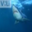}\includegraphics[width=0.4cm]{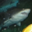}\includegraphics[width=0.4cm]{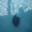}\includegraphics[width=0.4cm]{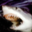}\includegraphics[width=0.4cm]{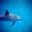}\includegraphics[width=0.4cm]{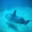}\includegraphics[width=0.4cm]{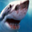}\includegraphics[width=0.4cm]{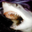}\includegraphics[width=0.4cm]{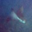}\includegraphics[width=0.4cm]{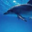}\includegraphics[width=0.4cm]{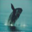}\includegraphics[width=0.4cm]{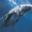}\includegraphics[width=0.4cm]{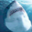}\includegraphics[width=0.4cm]{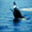}\includegraphics[width=0.4cm]{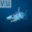}\includegraphics[width=0.4cm]{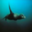}&fish and aquatic mammals\\
		\hline  5: bee,beetle,spider  & \includegraphics[width=0.4cm]{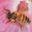}\includegraphics[width=0.4cm]{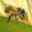}\includegraphics[width=0.4cm]{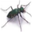}\includegraphics[width=0.4cm]{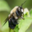}\includegraphics[width=0.4cm]{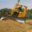}\includegraphics[width=0.4cm]{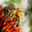}\includegraphics[width=0.4cm]{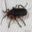}\includegraphics[width=0.4cm]{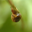}\includegraphics[width=0.4cm]{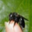}\includegraphics[width=0.4cm]{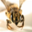}\includegraphics[width=0.4cm]{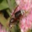}\includegraphics[width=0.4cm]{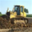}\includegraphics[width=0.4cm]{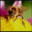}\includegraphics[width=0.4cm]{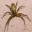}\includegraphics[width=0.4cm]{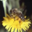}\includegraphics[width=0.4cm]{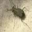}\includegraphics[width=0.4cm]{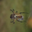}&insects\\
		\hline  6: tulip,rose,poppy  & \includegraphics[width=0.4cm]{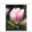}\includegraphics[width=0.4cm]{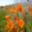}\includegraphics[width=0.4cm]{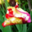}\includegraphics[width=0.4cm]{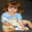}\includegraphics[width=0.4cm]{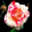}\includegraphics[width=0.4cm]{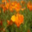}\includegraphics[width=0.4cm]{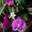}\includegraphics[width=0.4cm]{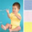}\includegraphics[width=0.4cm]{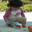}\includegraphics[width=0.4cm]{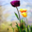}\includegraphics[width=0.4cm]{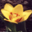}\includegraphics[width=0.4cm]{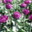}\includegraphics[width=0.4cm]{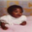}\includegraphics[width=0.4cm]{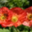}\includegraphics[width=0.4cm]{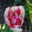}\includegraphics[width=0.4cm]{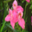}\includegraphics[width=0.4cm]{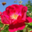}&flowers\\
		\hline  7: oak, willow, maple, pine  & \includegraphics[width=0.4cm]{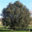}\includegraphics[width=0.4cm]{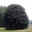}\includegraphics[width=0.4cm]{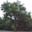}\includegraphics[width=0.4cm]{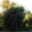}\includegraphics[width=0.4cm]{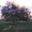}\includegraphics[width=0.4cm]{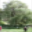}\includegraphics[width=0.4cm]{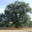}\includegraphics[width=0.4cm]{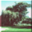}\includegraphics[width=0.4cm]{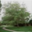}\includegraphics[width=0.4cm]{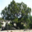}\includegraphics[width=0.4cm]{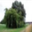}\includegraphics[width=0.4cm]{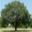}\includegraphics[width=0.4cm]{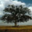}\includegraphics[width=0.4cm]{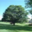}\includegraphics[width=0.4cm]{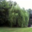}\includegraphics[width=0.4cm]{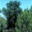}\includegraphics[width=0.4cm]{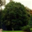}&trees\\
		\hline  8: telephone,cockroach,cup  & \includegraphics[width=0.4cm]{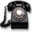}\includegraphics[width=0.4cm]{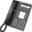}\includegraphics[width=0.4cm]{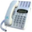}\includegraphics[width=0.4cm]{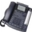}\includegraphics[width=0.4cm]{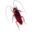}\includegraphics[width=0.4cm]{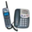}\includegraphics[width=0.4cm]{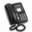}\includegraphics[width=0.4cm]{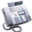}\includegraphics[width=0.4cm]{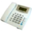}\includegraphics[width=0.4cm]{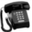}\includegraphics[width=0.4cm]{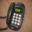}\includegraphics[width=0.4cm]{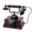}\includegraphics[width=0.4cm]{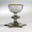}\includegraphics[width=0.4cm]{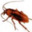}\includegraphics[width=0.4cm]{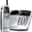}\includegraphics[width=0.4cm]{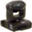}\includegraphics[width=0.4cm]{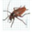}&household electrical devices\\
		\hline  9: hamster,cockroach,mouse  & \includegraphics[width=0.4cm]{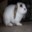}\includegraphics[width=0.4cm]{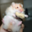}\includegraphics[width=0.4cm]{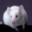}\includegraphics[width=0.4cm]{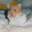}\includegraphics[width=0.4cm]{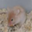}\includegraphics[width=0.4cm]{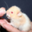}\includegraphics[width=0.4cm]{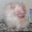}\includegraphics[width=0.4cm]{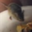}\includegraphics[width=0.4cm]{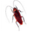}\includegraphics[width=0.4cm]{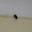}\includegraphics[width=0.4cm]{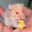}\includegraphics[width=0.4cm]{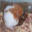}\includegraphics[width=0.4cm]{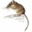}\includegraphics[width=0.4cm]{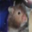}\includegraphics[width=0.4cm]{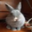}\includegraphics[width=0.4cm]{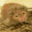}\includegraphics[width=0.4cm]{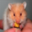}&small mammals\\
		
		\hline  10: boy,woman,girl,baby  & \includegraphics[width=0.4cm]{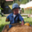}\includegraphics[width=0.4cm]{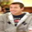}\includegraphics[width=0.4cm]{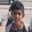}\includegraphics[width=0.4cm]{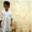}\includegraphics[width=0.4cm]{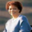}\includegraphics[width=0.4cm]{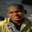}\includegraphics[width=0.4cm]{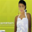}\includegraphics[width=0.4cm]{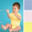}\includegraphics[width=0.4cm]{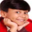}\includegraphics[width=0.4cm]{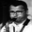}\includegraphics[width=0.4cm]{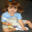}\includegraphics[width=0.4cm]{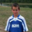}\includegraphics[width=0.4cm]{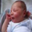}\includegraphics[width=0.4cm]{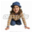}\includegraphics[width=0.4cm]{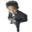}\includegraphics[width=0.4cm]{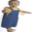}\includegraphics[width=0.4cm]{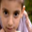}&people\\
		\hline  11: aquarium fish,trout  & \includegraphics[width=0.4cm]{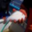}\includegraphics[width=0.4cm]{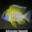}\includegraphics[width=0.4cm]{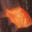}\includegraphics[width=0.4cm]{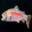}\includegraphics[width=0.4cm]{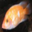}\includegraphics[width=0.4cm]{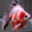}\includegraphics[width=0.4cm]{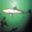}\includegraphics[width=0.4cm]{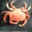}\includegraphics[width=0.4cm]{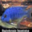}\includegraphics[width=0.4cm]{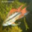}\includegraphics[width=0.4cm]{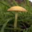}\includegraphics[width=0.4cm]{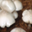}\includegraphics[width=0.4cm]{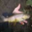}\includegraphics[width=0.4cm]{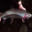}\includegraphics[width=0.4cm]{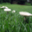}\includegraphics[width=0.4cm]{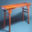}\includegraphics[width=0.4cm]{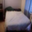}&fish\\
		\hline  12: lawn mover,camel  & \includegraphics[width=0.4cm]{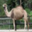}\includegraphics[width=0.4cm]{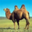}\includegraphics[width=0.4cm]{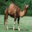}\includegraphics[width=0.4cm]{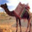}\includegraphics[width=0.4cm]{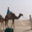}\includegraphics[width=0.4cm]{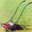}\includegraphics[width=0.4cm]{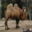}\includegraphics[width=0.4cm]{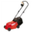}\includegraphics[width=0.4cm]{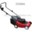}\includegraphics[width=0.4cm]{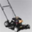}\includegraphics[width=0.4cm]{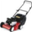}\includegraphics[width=0.4cm]{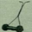}\includegraphics[width=0.4cm]{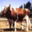}\includegraphics[width=0.4cm]{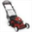}\includegraphics[width=0.4cm]{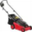}\includegraphics[width=0.4cm]{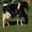}\includegraphics[width=0.4cm]{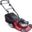}&large omnivores and herbivores\\
		\hline  13: motorcycle,bicycle,tiger  & \includegraphics[width=0.4cm]{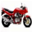}\includegraphics[width=0.4cm]{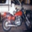}\includegraphics[width=0.4cm]{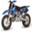}\includegraphics[width=0.4cm]{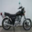}\includegraphics[width=0.4cm]{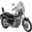}\includegraphics[width=0.4cm]{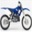}\includegraphics[width=0.4cm]{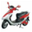}\includegraphics[width=0.4cm]{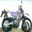}\includegraphics[width=0.4cm]{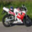}\includegraphics[width=0.4cm]{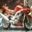}\includegraphics[width=0.4cm]{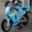}\includegraphics[width=0.4cm]{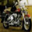}\includegraphics[width=0.4cm]{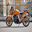}\includegraphics[width=0.4cm]{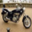}\includegraphics[width=0.4cm]{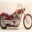}\includegraphics[width=0.4cm]{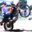}\includegraphics[width=0.4cm]{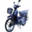}&vehicles1\\
		\hline  14: sea,plain,could,mountain  & \includegraphics[width=0.4cm]{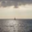}\includegraphics[width=0.4cm]{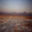}\includegraphics[width=0.4cm]{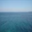}\includegraphics[width=0.4cm]{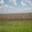}\includegraphics[width=0.4cm]{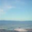}\includegraphics[width=0.4cm]{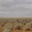}\includegraphics[width=0.4cm]{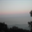}\includegraphics[width=0.4cm]{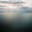}\includegraphics[width=0.4cm]{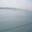}\includegraphics[width=0.4cm]{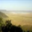}\includegraphics[width=0.4cm]{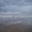}\includegraphics[width=0.4cm]{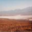}\includegraphics[width=0.4cm]{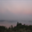}\includegraphics[width=0.4cm]{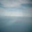}\includegraphics[width=0.4cm]{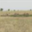}\includegraphics[width=0.4cm]{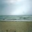}\includegraphics[width=0.4cm]{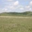}&large natural outdoor scenes\\
		\hline  15: caterpillar,skunk,worm  & \includegraphics[width=0.4cm]{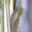}\includegraphics[width=0.4cm]{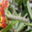}\includegraphics[width=0.4cm]{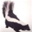}\includegraphics[width=0.4cm]{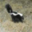}\includegraphics[width=0.4cm]{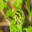}\includegraphics[width=0.4cm]{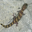}\includegraphics[width=0.4cm]{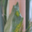}\includegraphics[width=0.4cm]{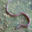}\includegraphics[width=0.4cm]{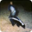}\includegraphics[width=0.4cm]{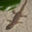}\includegraphics[width=0.4cm]{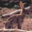}\includegraphics[width=0.4cm]{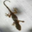}\includegraphics[width=0.4cm]{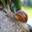}\includegraphics[width=0.4cm]{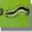}\includegraphics[width=0.4cm]{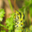}\includegraphics[width=0.4cm]{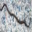}\includegraphics[width=0.4cm]{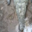}&reptiles\\
		\hline  16: apple,orange,pear  & \includegraphics[width=0.4cm]{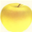}\includegraphics[width=0.4cm]{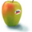}\includegraphics[width=0.4cm]{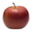}\includegraphics[width=0.4cm]{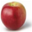}\includegraphics[width=0.4cm]{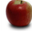}\includegraphics[width=0.4cm]{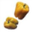}\includegraphics[width=0.4cm]{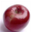}\includegraphics[width=0.4cm]{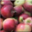}\includegraphics[width=0.4cm]{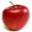}\includegraphics[width=0.4cm]{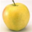}\includegraphics[width=0.4cm]{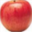}\includegraphics[width=0.4cm]{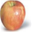}\includegraphics[width=0.4cm]{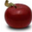}\includegraphics[width=0.4cm]{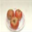}\includegraphics[width=0.4cm]{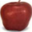}\includegraphics[width=0.4cm]{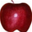}\includegraphics[width=0.4cm]{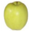}&fruit and vegetables\\
		\hline  17: hamster,raccoon,wolf  & \includegraphics[width=0.4cm]{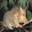}\includegraphics[width=0.4cm]{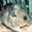}\includegraphics[width=0.4cm]{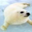}\includegraphics[width=0.4cm]{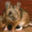}\includegraphics[width=0.4cm]{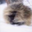}\includegraphics[width=0.4cm]{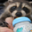}\includegraphics[width=0.4cm]{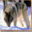}\includegraphics[width=0.4cm]{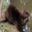}\includegraphics[width=0.4cm]{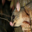}\includegraphics[width=0.4cm]{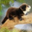}\includegraphics[width=0.4cm]{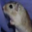}\includegraphics[width=0.4cm]{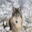}\includegraphics[width=0.4cm]{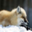}\includegraphics[width=0.4cm]{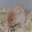}\includegraphics[width=0.4cm]{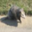}\includegraphics[width=0.4cm]{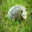}\includegraphics[width=0.4cm]{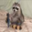}&medium mammals\\
		\hline  18: castle,house,wardrobe  & \includegraphics[width=0.4cm]{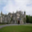}\includegraphics[width=0.4cm]{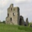}\includegraphics[width=0.4cm]{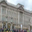}\includegraphics[width=0.4cm]{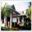}\includegraphics[width=0.4cm]{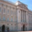}\includegraphics[width=0.4cm]{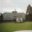}\includegraphics[width=0.4cm]{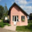}\includegraphics[width=0.4cm]{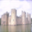}\includegraphics[width=0.4cm]{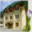}\includegraphics[width=0.4cm]{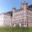}\includegraphics[width=0.4cm]{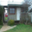}\includegraphics[width=0.4cm]{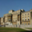}\includegraphics[width=0.4cm]{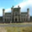}\includegraphics[width=0.4cm]{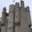}\includegraphics[width=0.4cm]{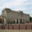}\includegraphics[width=0.4cm]{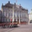}\includegraphics[width=0.4cm]{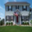}&large man made outdoor things\\
		\hline  19: plate,cup,bowl  & \includegraphics[width=0.4cm]{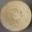}\includegraphics[width=0.4cm]{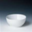}\includegraphics[width=0.4cm]{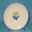}\includegraphics[width=0.4cm]{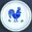}\includegraphics[width=0.4cm]{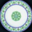}\includegraphics[width=0.4cm]{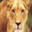}\includegraphics[width=0.4cm]{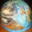}\includegraphics[width=0.4cm]{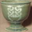}\includegraphics[width=0.4cm]{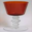}\includegraphics[width=0.4cm]{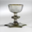}\includegraphics[width=0.4cm]{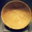}\includegraphics[width=0.4cm]{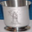}\includegraphics[width=0.4cm]{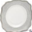}\includegraphics[width=0.4cm]{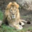}\includegraphics[width=0.4cm]{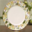}\includegraphics[width=0.4cm]{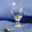}\includegraphics[width=0.4cm]{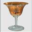}&food containers\\
		\hline
		
	\end{tabular}
\end{table*}
\subsubsection{Layer-wise Analysis of Neuron representations:}
\label{sec:activations}
\begin{figure*}[t!]
	\begin{subfigure}{0.3\textwidth}
		\centering
		\includegraphics[width=\linewidth]{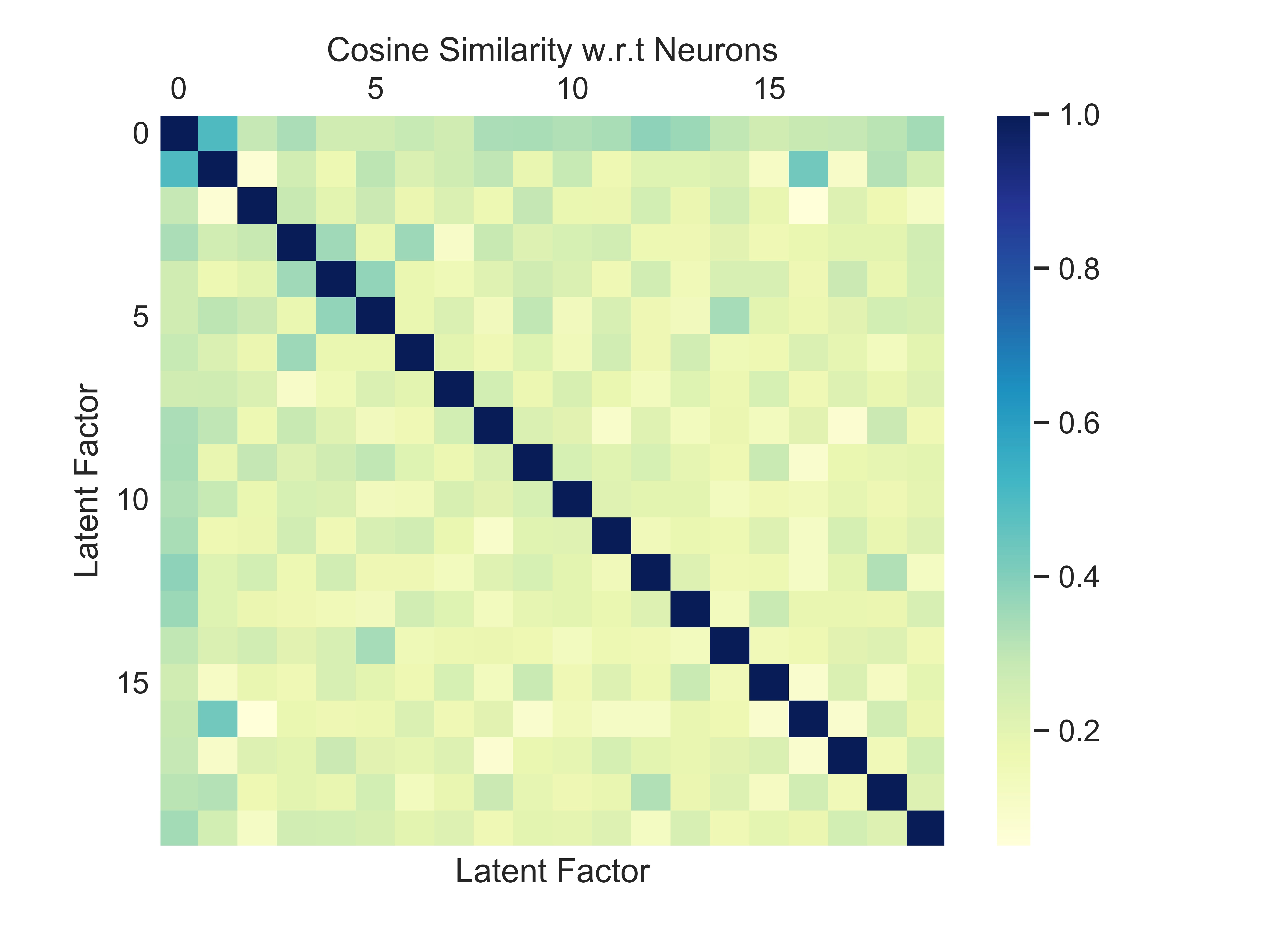}
		\caption{Cosine Similarity: Layer 0}
		\label{fig:resnetcos0}
	\end{subfigure}
	\begin{subfigure}{0.3\textwidth}
		\centering
		\includegraphics[width=\linewidth]{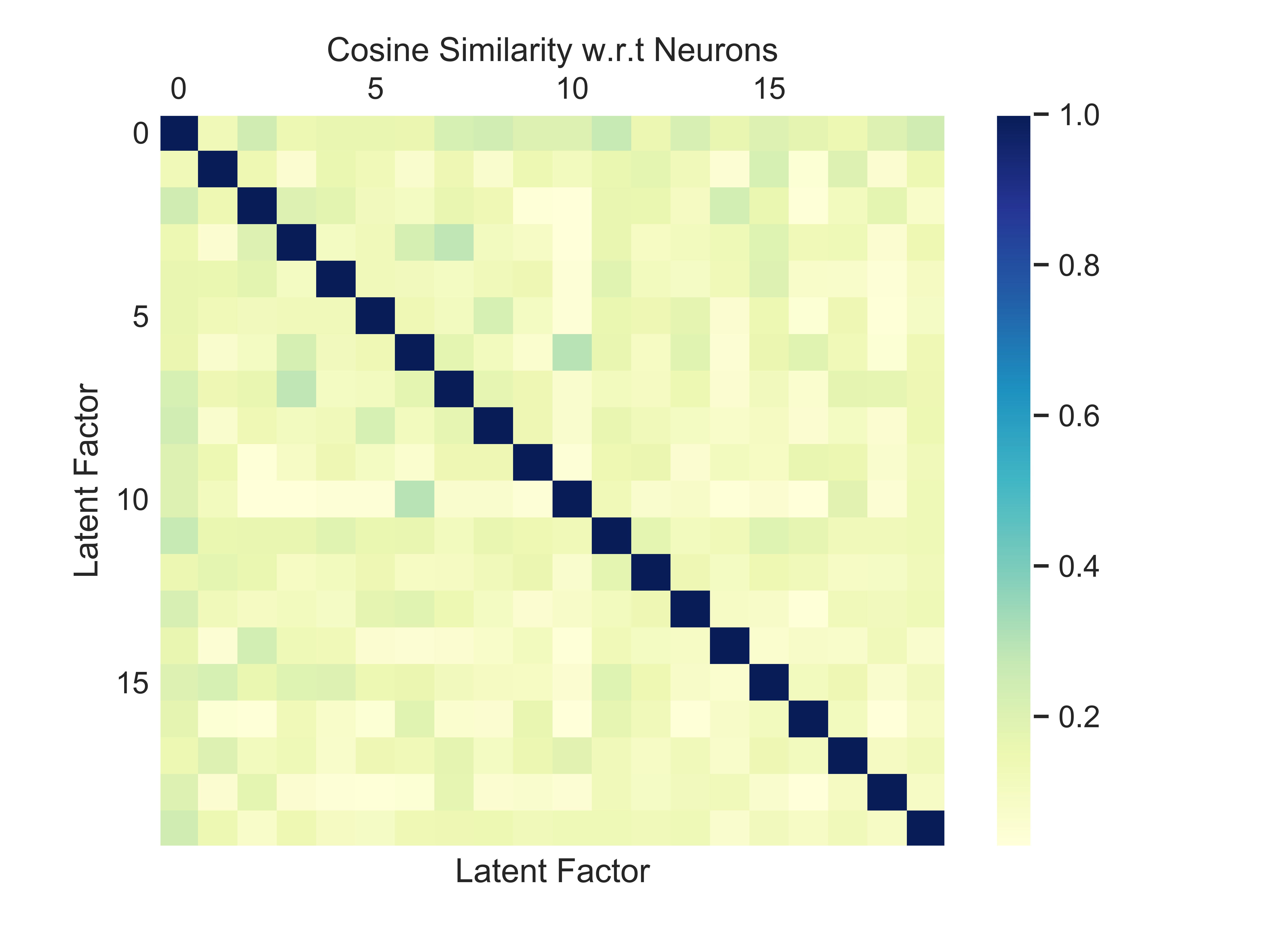}
		\caption{Cosine Similarity: Layer 1}
		\label{fig:resnetcos1}
	\end{subfigure}
	\begin{subfigure}{0.3\textwidth}
		\centering
		\includegraphics[width=\linewidth]{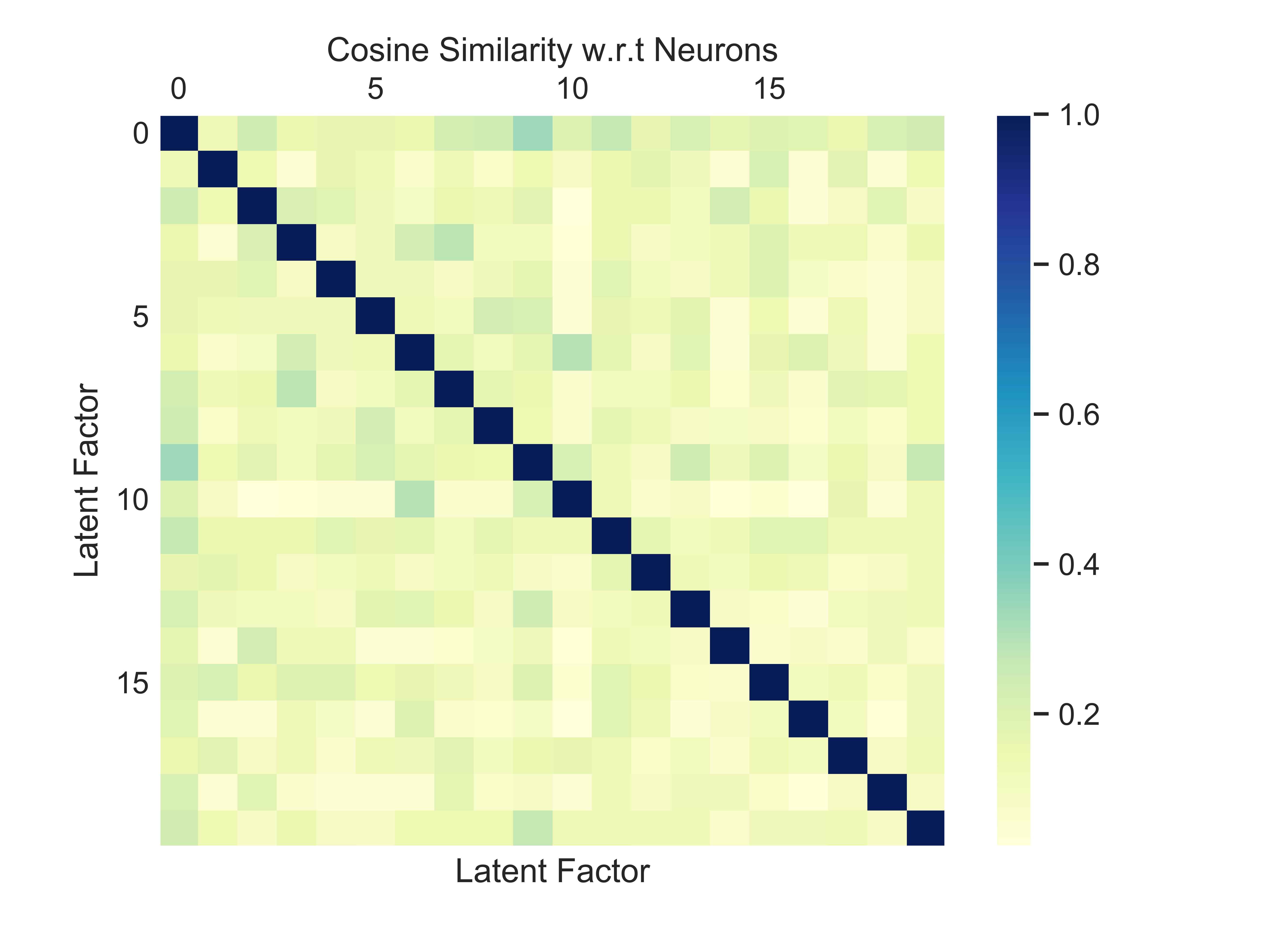}
		\caption{Cosine Similarity: Layer 2}
		\label{fig:resnetcos2}
	\end{subfigure}
	\caption{Plots of Cosine Similarity of Latent Factors in Layers 0,1,2 of ResNet-18. This highlighits the layerwise learning dynamics of the network and helps us visualize with concepts and classes occupy similar neural regions in a given layer of a network and how they evolve as we go deeper into the network. In fact, we observe that as we go deeper into the network, the similarity bcomes diagonal, showing higher separation of the latent concepts}
	\label{fig:figResnet1b}
\end{figure*}

\begin{figure*}[thb!]
	\begin{subfigure}{0.5\textwidth}
	\centering
	\includegraphics[width=0.8\linewidth]{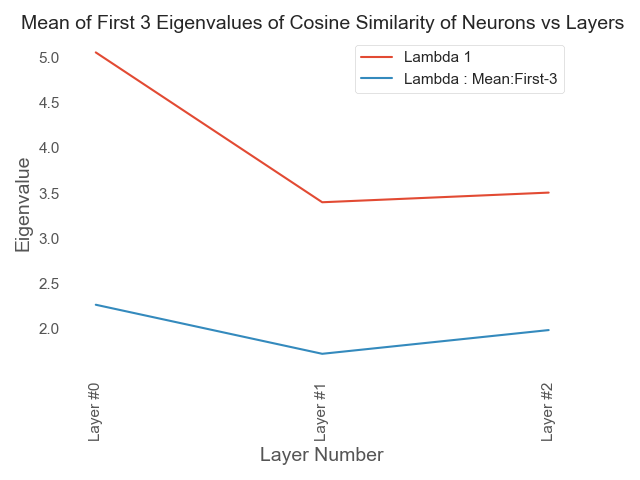}
	\caption{Eigenvalues of Similarity Matrices of ResNet-18}
	\label{fig:resneteig}
	\end{subfigure}\hfill
	\begin{subfigure}{0.5\textwidth}
		\centering
		\includegraphics[width=0.8\linewidth]{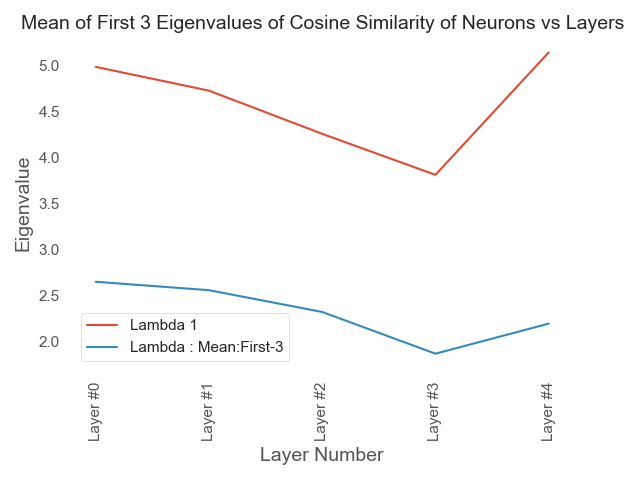}
		\caption{Eigenvalues of Similarity Matrices of VGG-11}
		\label{fig:vgg11eig}
	\end{subfigure}
	\caption{Plots of Eigenvalues of Similarity Matrix for ResNet-18 and VGG-11. This plot shows increasing independence of learned latent concepts w.r.t. neurons as we go deeper in the non-classification layers of the network. The closer the a matrix is to Identity the closer the average of it's eigen values is to 1 and vice versa. The last layer in each of the 2 figures is the output of pre log softmax of the network, which is usually a much lower dimensional space than the previous layers}
	\label{fig:resnet18vgg11eig}
\end{figure*}

In this section we try to quantify the behaviour of neurons as a cluster and across layers. We utilize the neuron embedding matrix for a given layer $j$, as denoted by $O_j$ $\in$ $ \mathbb{R}^{N_j \times d}_{+}$, where $N_j$ is the number of neurons in layer $j$, whereas $d$ is the number of latent factors in the factorization. Next we compute pairwise cosine similarity between the columns of a matrix $O_j$ and we do this $\forall j$ as shown in Figure \ref{fig:resnetcos0}, Figure \ref{fig:resnetcos1} and Figure \ref{fig:resnetcos2}. Here Layer 0,1,2 refer to 3 layers analyzed in the ResNet-18 in increasing order of depth and are not necessarily the first,second and third layers of the network. In these plots a high value at any entry $(i,j)$ indicates a higher overlap between the number of neurons which fire for inputs belonging in the 2 super classes best approximated by latent factor $i$ and latent factor $j$. 
As indicated in Figures \ref{fig:resnetcos0},\ref{fig:resnetcos1} and \ref{fig:resnetcos2} the activations tend to be more intra-superclass, a result similar in nature to one observed by SVCCA\cite{raghu2017svcca} , i.e. more concentrated along the diagonal of the Similarity Matrix as we go deeper down the layers. This is also borne out by the eigen values of these Similarity Matrices, as the matrices tend to get closer to Identity, the lower the mean of first-K eigen-values as shown in \ref{fig:resneteig}.

We also show similar results for a VGG-11\cite{simonyan2014deep} Trained and analyzed on CIFAR-100 in Figure \ref{fig:vgg11eig}.




\subsubsection{Co-Analysis of Pixels and Inputs:}
\label{sec:pixels}
\begin{figure} [!htb]
	\begin{subfigure}[b]{0.3\linewidth}
		\centering
		\includegraphics[width=0.9\textwidth]{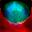}
		\label{fig:lf2a}
		\caption{Latent-Factor:2}
	\end{subfigure}
	\begin{subfigure}[b]{0.3\linewidth}    
		\centering
		\includegraphics[width=0.9\textwidth]{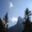}
		\label{fig:lf2b}
		\caption{Image:Mountain}    
	\end{subfigure}
	\begin{subfigure}[b]{0.3\linewidth}
		\centering    
		\includegraphics[width=0.9\textwidth]{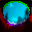}
		\label{fig:lf2c}
		\caption{Filtered Image}
	\end{subfigure}
	\begin{subfigure}{0.4\textwidth}
		\centering
		\includegraphics[width=0.8\linewidth]{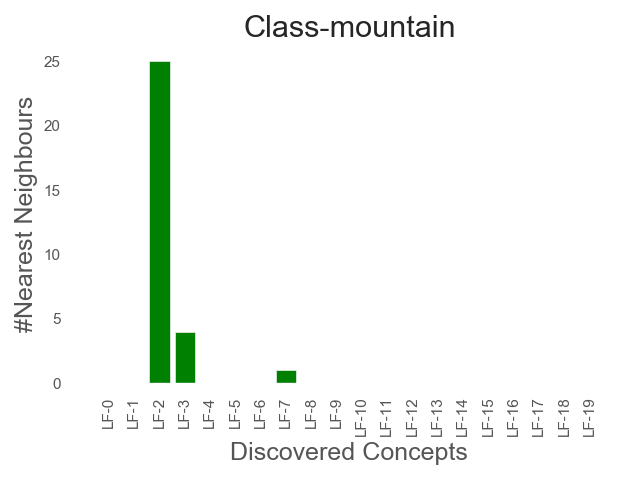}
		\caption{Top-30 Nearest Neighbours of this Image}
		\label{fig:lf2d}
	\end{subfigure}
	\caption{Analysis of Topmost Image from Latent-Factor:2}
	\label{fig:LF2}
\end{figure}

\begin{figure} [!htb]
	\begin{subfigure}[b]{0.3\linewidth}
		\centering
		\includegraphics[width=0.9\textwidth]{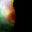}
		\label{fig:3a}
		\caption{Latent-Factor:3}
	\end{subfigure}
	\begin{subfigure}[b]{0.3\linewidth}    
		\centering
		\includegraphics[width=0.9\textwidth]{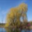}
		\label{fig:3b}
		\caption{Image: Willow-Tree}    
	\end{subfigure}
	\begin{subfigure}[b]{0.3\linewidth}
		\centering    
		\includegraphics[width=0.9\textwidth]{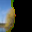}
		\label{fig:3c}
		\caption{Filtered Image}
	\end{subfigure}
	\begin{subfigure}{0.4\textwidth}
		\centering
		\includegraphics[width=0.8\linewidth]{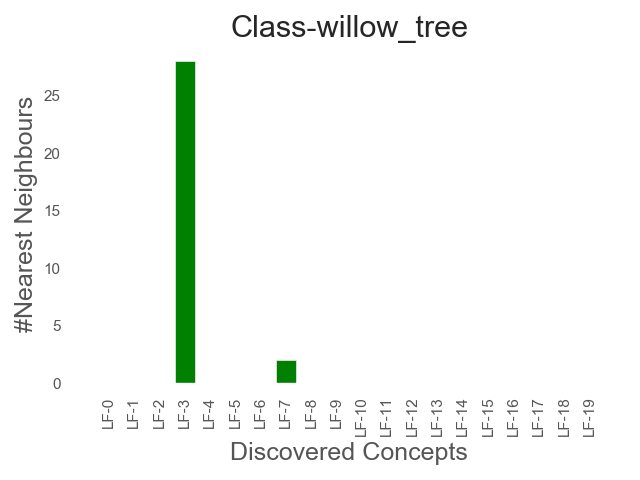}
		\caption{Top-30 Nearest Neighbours of this Image}
		\label{fig:3d}
	\end{subfigure}
	\caption{Analysis of Topmost Image from Latent-Factor:3}
	\label{fig:LF3}
\end{figure}

In this section we analyze the pixel space along with inputs. The Matrices $P_i$'s $\in$ $ \mathbb{R}^{S_i \times d}_{+}$ hold the input representation of pixels in the input channel $i$ where $S_i$ is the number of Pixels in Input Channel-$i$, or the vectorized size of the channel. Each column of a matrix $P_i$ represents a feature activation score of all the pixels in channel $i$ for the given latent factor. Therefore by collecting information from column 2 of $P_0$,$P_1$ and $P_2$ and resizing them appropriately we get an average pattern of activation across the pixel space for all the images that belong to Latent Factor 2, as shown in Figure \ref{fig:LF2}a, and for Latent-Factor-3 in \ref{fig:LF3}a. This functionality is very similar to LIME \cite{lime}, but instead of individual images we can operate on pixel representations which represent learned concepts. We then take these Latent-Factor-Images, and create a mask where we assign a value of 1 at a pixel location if it's activation value is above the median activation value for the Latent Image and 0 other wise and overlay it with the topmost images of the Latent Factor as found in our analysis of Matrix-$F$ in Table \ref{tab:Table1}. We also take around 30 Nearest Neighbours of the Image as determined by the Latent Space of Matrix-$F$ and give a distribution of the Latent Concepts those Neighbours have their highest affinity for, thereby helping us achieve interpretability on an input-by-input basis by being able to say that a given image is close to another. Next, via 2 examples we present a per example case study of interpretability possible by the use of this model.\\
In Figures \ref{fig:LF2}a,\ref{fig:LF2}b,\ref{fig:LF2}c and \ref{fig:LF2}d For Latent Factor-2 we present the Latent Representations of Pixels, The topmost Image in that Latent Factor, The top 50\% activated pixels super imposed on the original image, and the Latent Concept Distribution of top-30 Nearest Neighbours of the image, respectively. As noted previously in Table\ref{tab:Table1}, Latent Factor 2 Represents classes like mountain, bridge, castles, skyscrapers etc, leading to it's topmost super class being "large man-made outdoor things". On average, the most activated pixels for images belonging to this superclass tend to be blue pixels towards the top, green towards the middle and red towards the bottom. And the set of top-30 nearest neighbours for this particular image of a Mountain also has members belonging to Latent Factor 3 and 7, 2 concepts which have a high affinity for inputs belonging to super class of trees.\\
In Figures \ref{fig:LF3}a,\ref{fig:LF3}b,\ref{fig:LF3}c and \ref{fig:LF3}d we present a similar analysis for Latent Factor-3. As shown in Table\ref{tab:Table1}, Latent Factor 3 Represents classes like willow tree, maple tree , oak trees, pine trees etc, leading to it's topmost super class being "trees". In this case, the most activated set of pixels is on the right half of the pixel space with a higher affinity in green and red channels of the image, as shown in Figure \ref{fig:LF3}a. In Figure \ref{fig:LF3}c we see the effects of applying this latent image as a filter to the image of a willow tree \ref{fig:LF3}b and we see that the right half of the image is redundant and the left half captures basic underlying features about the image like contours, shapes colours etc.


\section{Extensions and Applications of the Model}
\label{sec:ext}
\begin{figure*}[!h]
		\centering
		\includegraphics[width=\linewidth]{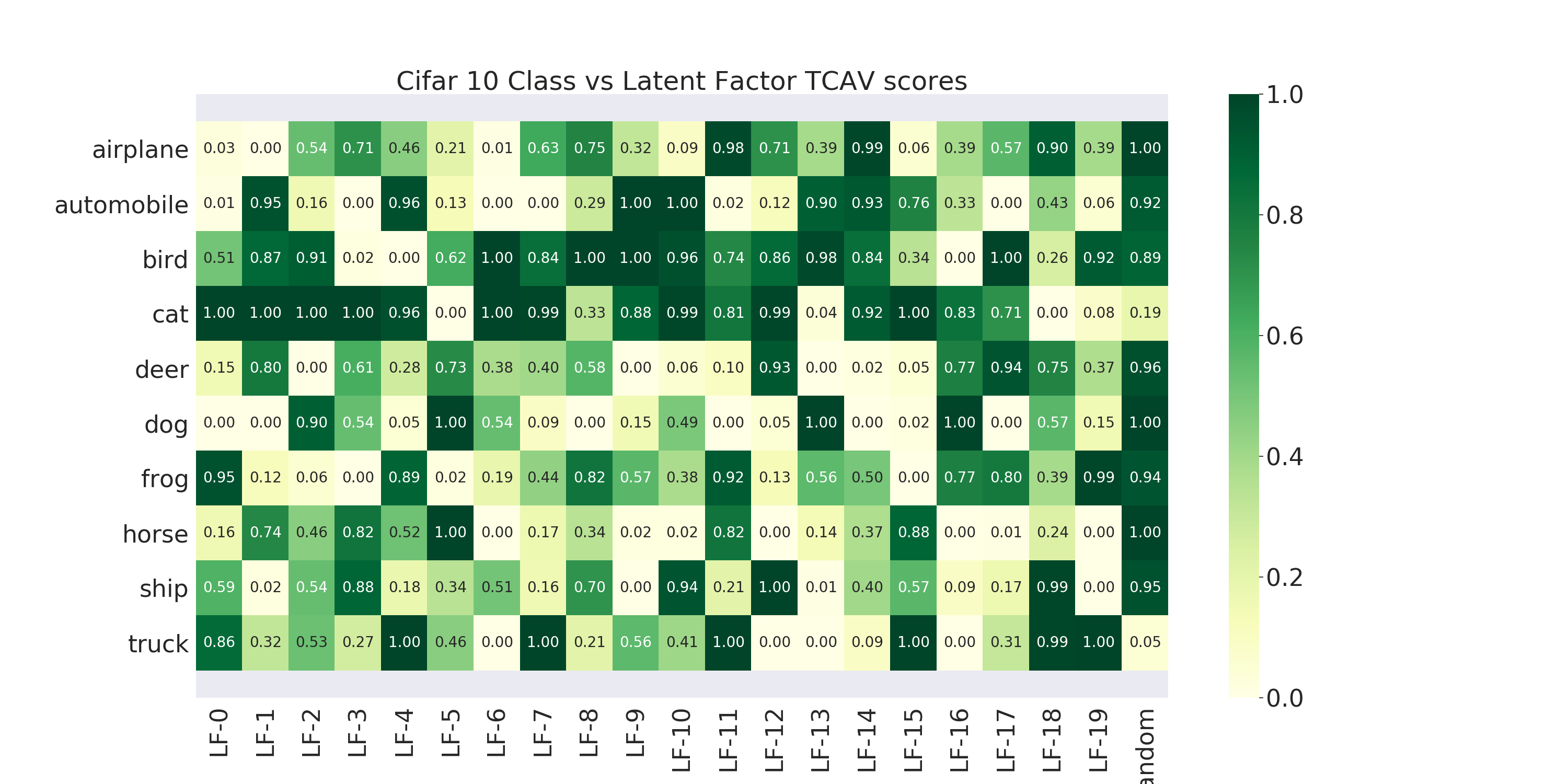}
		\caption{TCAV scores for Group Sparsity based model}
	\label{fig:threePlusfour}
\end{figure*}

We modify the model in \autoref{eq:1} by imposing a group sparse regularization \cite{doi:10.1137/1.9781611972825.73} on the factor Matrix-$F$ and only including the neural activations in the objective function. In \autoref{tab:Table2}, we present a case study where a ResNet-18 is trained on CIFAR-10 \cite{Krizhevsky09learningmultiple} and evaluated on CIFAR-100, we omit some latent factors for brevity. The goal here is to visualize the generalization ability of the network to cluster and distinguish between natural images based on an observation of similar but out of sample data distribution during training. In order to validate the result we then take the latent concepts learned by the model and evaluate TCAV\cite{kim2017interpretability} scores \footnote{https://github.com/rakhimovv/tcav}  for each combination of latent concept and input class, the results for which are shown in \autoref{fig:threePlusfour}.
\begin{table*}[!ht]
	\caption{Latent Factor Analysis Group Sparsity - Activations Only} \label{tab:Table2}
	\begin{tabular}
		{lllll} \hline Factor: Top Classes  & Top Images\\
		
		\hline   1: kangaroo,rabbit,fox,squirrel  & \includegraphics[width=0.4cm]{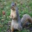}\includegraphics[width=0.4cm]{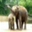}\includegraphics[width=0.4cm]{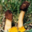}\includegraphics[width=0.4cm]{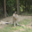}\includegraphics[width=0.4cm]{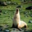}\includegraphics[width=0.4cm]{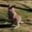}\includegraphics[width=0.4cm]{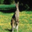}\includegraphics[width=0.4cm]{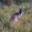}\includegraphics[width=0.4cm]{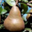}\includegraphics[width=0.4cm]{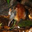}\includegraphics[width=0.4cm]{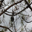}\includegraphics[width=0.4cm]{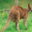}\includegraphics[width=0.4cm]{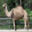}\includegraphics[width=0.4cm]{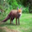}\includegraphics[width=0.4cm]{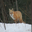}\includegraphics[width=0.4cm]{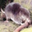}\includegraphics[width=0.4cm]{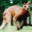}\includegraphics[width=0.4cm]{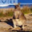}\includegraphics[width=0.4cm]{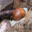}\includegraphics[width=0.4cm]{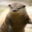}\\
		\hline   2: woman,boy,girl,baby,man  & \includegraphics[width=0.4cm]{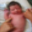}\includegraphics[width=0.4cm]{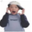}\includegraphics[width=0.4cm]{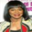}\includegraphics[width=0.4cm]{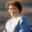}\includegraphics[width=0.4cm]{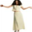}\includegraphics[width=0.4cm]{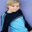}\includegraphics[width=0.4cm]{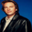}\includegraphics[width=0.4cm]{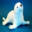}\includegraphics[width=0.4cm]{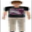}\includegraphics[width=0.4cm]{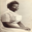}\includegraphics[width=0.4cm]{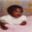}\includegraphics[width=0.4cm]{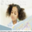}\includegraphics[width=0.4cm]{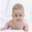}\includegraphics[width=0.4cm]{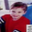}\includegraphics[width=0.4cm]{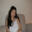}\includegraphics[width=0.4cm]{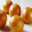}\includegraphics[width=0.4cm]{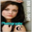}\includegraphics[width=0.4cm]{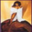}\includegraphics[width=0.4cm]{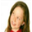}\includegraphics[width=0.4cm]{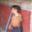}\\
		\hline   4: pickup-truck,motorcycle,bus,tank  & \includegraphics[width=0.4cm]{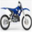}\includegraphics[width=0.4cm]{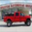}\includegraphics[width=0.4cm]{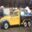}\includegraphics[width=0.4cm]{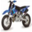}\includegraphics[width=0.4cm]{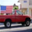}\includegraphics[width=0.4cm]{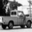}\includegraphics[width=0.4cm]{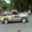}\includegraphics[width=0.4cm]{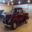}\includegraphics[width=0.4cm]{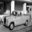}\includegraphics[width=0.4cm]{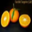}\includegraphics[width=0.4cm]{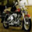}\includegraphics[width=0.4cm]{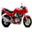}\includegraphics[width=0.4cm]{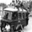}\includegraphics[width=0.4cm]{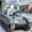}\includegraphics[width=0.4cm]{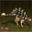}\includegraphics[width=0.4cm]{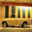}\includegraphics[width=0.4cm]{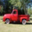}\includegraphics[width=0.4cm]{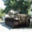}\includegraphics[width=0.4cm]{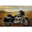}\includegraphics[width=0.4cm]{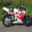}\\
		\hline   5: cattle,elephant,camel,chimpanzee  & \includegraphics[width=0.4cm]{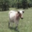}\includegraphics[width=0.4cm]{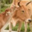}\includegraphics[width=0.4cm]{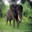}\includegraphics[width=0.4cm]{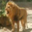}\includegraphics[width=0.4cm]{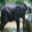}\includegraphics[width=0.4cm]{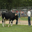}\includegraphics[width=0.4cm]{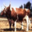}\includegraphics[width=0.4cm]{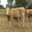}\includegraphics[width=0.4cm]{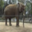}\includegraphics[width=0.4cm]{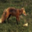}\includegraphics[width=0.4cm]{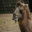}\includegraphics[width=0.4cm]{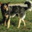}\includegraphics[width=0.4cm]{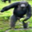}\includegraphics[width=0.4cm]{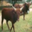}\includegraphics[width=0.4cm]{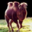}\includegraphics[width=0.4cm]{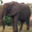}\includegraphics[width=0.4cm]{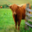}\includegraphics[width=0.4cm]{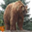}\includegraphics[width=0.4cm]{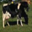}\includegraphics[width=0.4cm]{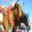}\\
		\hline  6: porcupine,possum,squirrel,raccoon  & \includegraphics[width=0.4cm]{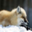}\includegraphics[width=0.4cm]{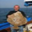}
		\includegraphics[width=0.4cm]{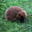}\includegraphics[width=0.4cm]{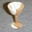}\includegraphics[width=0.4cm]{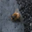}\includegraphics[width=0.4cm]{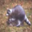}\includegraphics[width=0.4cm]{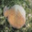}\includegraphics[width=0.4cm]{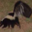}\includegraphics[width=0.4cm]{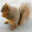}\includegraphics[width=0.4cm]{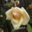}\includegraphics[width=0.4cm]{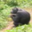}\includegraphics[width=0.4cm]{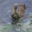}\includegraphics[width=0.4cm]{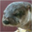}\includegraphics[width=0.4cm]{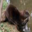}\includegraphics[width=0.4cm]{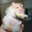}\includegraphics[width=0.4cm]{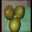}\includegraphics[width=0.4cm]{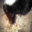}\includegraphics[width=0.4cm]{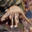}\includegraphics[width=0.4cm]{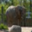}\includegraphics[width=0.4cm]{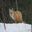}\\
		\hline   7: willow,maple,oak,palm -all trees  & \includegraphics[width=0.4cm]{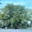}\includegraphics[width=0.4cm]{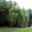}\includegraphics[width=0.4cm]{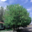}\includegraphics[width=0.4cm]{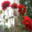}\includegraphics[width=0.4cm]{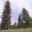}\includegraphics[width=0.4cm]{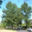}\includegraphics[width=0.4cm]{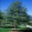}\includegraphics[width=0.4cm]{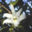}\includegraphics[width=0.4cm]{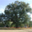}\includegraphics[width=0.4cm]{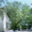}\includegraphics[width=0.4cm]{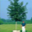}\includegraphics[width=0.4cm]{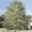}\includegraphics[width=0.4cm]{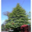}\includegraphics[width=0.4cm]{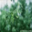}\includegraphics[width=0.4cm]{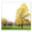}\includegraphics[width=0.4cm]{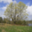}\includegraphics[width=0.4cm]{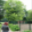}\includegraphics[width=0.4cm]{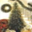}\includegraphics[width=0.4cm]{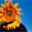}\includegraphics[width=0.4cm]{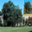}\\
		\hline   9: apple,sweet-pepper,rose,orange  & \includegraphics[width=0.4cm]{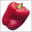}\includegraphics[width=0.4cm]{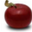}\includegraphics[width=0.4cm]{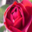}\includegraphics[width=0.4cm]{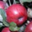}\includegraphics[width=0.4cm]{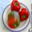}\includegraphics[width=0.4cm]{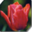}\includegraphics[width=0.4cm]{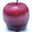}\includegraphics[width=0.4cm]{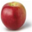}\includegraphics[width=0.4cm]{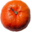}\includegraphics[width=0.4cm]{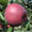}\includegraphics[width=0.4cm]{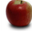}\includegraphics[width=0.4cm]{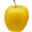}\includegraphics[width=0.4cm]{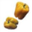}\includegraphics[width=0.4cm]{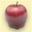}\includegraphics[width=0.4cm]{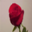}\includegraphics[width=0.4cm]{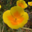}\includegraphics[width=0.4cm]{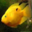}\includegraphics[width=0.4cm]{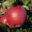}\includegraphics[width=0.4cm]{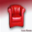}\includegraphics[width=0.4cm]{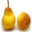}\\
		\hline   10: plate,bowl,can,clock  & \includegraphics[width=0.4cm]{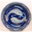}\includegraphics[width=0.4cm]{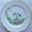}\includegraphics[width=0.4cm]{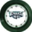}\includegraphics[width=0.4cm]{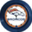}\includegraphics[width=0.4cm]{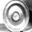}\includegraphics[width=0.4cm]{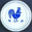}\includegraphics[width=0.4cm]{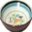}\includegraphics[width=0.4cm]{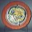}\includegraphics[width=0.4cm]{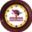}\includegraphics[width=0.4cm]{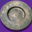}\includegraphics[width=0.4cm]{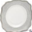}\includegraphics[width=0.4cm]{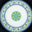}\includegraphics[width=0.4cm]{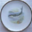}\includegraphics[width=0.4cm]{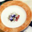}\includegraphics[width=0.4cm]{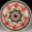}\includegraphics[width=0.4cm]{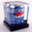}\includegraphics[width=0.4cm]{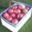}\includegraphics[width=0.4cm]{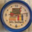}\includegraphics[width=0.4cm]{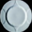}\includegraphics[width=0.4cm]{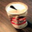}\\
		\hline   12: whale,rocket,dolphin,sea  & \includegraphics[width=0.4cm]{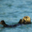}\includegraphics[width=0.4cm]{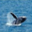}\includegraphics[width=0.4cm]{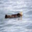}\includegraphics[width=0.4cm]{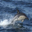}\includegraphics[width=0.4cm]{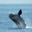}\includegraphics[width=0.4cm]{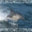}\includegraphics[width=0.4cm]{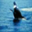}\includegraphics[width=0.4cm]{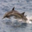}\includegraphics[width=0.4cm]{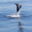}\includegraphics[width=0.4cm]{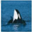}\includegraphics[width=0.4cm]{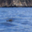}\includegraphics[width=0.4cm]{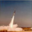}\includegraphics[width=0.4cm]{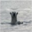}\includegraphics[width=0.4cm]{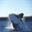}\includegraphics[width=0.4cm]{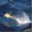}\includegraphics[width=0.4cm]{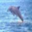}\includegraphics[width=0.4cm]{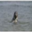}\includegraphics[width=0.4cm]{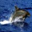}\includegraphics[width=0.4cm]{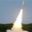}\includegraphics[width=0.4cm]{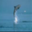}\\
		\hline   13: girl,snail,boy,spider,crab  & \includegraphics[width=0.4cm]{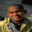}\includegraphics[width=0.4cm]{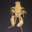}\includegraphics[width=0.4cm]{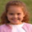}\includegraphics[width=0.4cm]{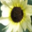}\includegraphics[width=0.4cm]{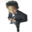}\includegraphics[width=0.4cm]{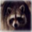}\includegraphics[width=0.4cm]{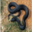}\includegraphics[width=0.4cm]{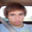}\includegraphics[width=0.4cm]{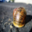}\includegraphics[width=0.4cm]{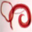}\includegraphics[width=0.4cm]{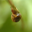}\includegraphics[width=0.4cm]{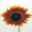}\includegraphics[width=0.4cm]{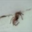}\includegraphics[width=0.4cm]{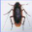}\includegraphics[width=0.4cm]{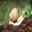}\includegraphics[width=0.4cm]{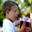}\includegraphics[width=0.4cm]{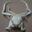}\includegraphics[width=0.4cm]{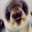}\includegraphics[width=0.4cm]{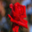}\includegraphics[width=0.4cm]{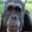}\\
		\hline   16: lion,hamster,wolf,mouse  & \includegraphics[width=0.4cm]{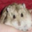}\includegraphics[width=0.4cm]{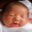}\includegraphics[width=0.4cm]{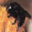}\includegraphics[width=0.4cm]{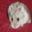}\includegraphics[width=0.4cm]{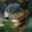}\includegraphics[width=0.4cm]{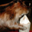}\includegraphics[width=0.4cm]{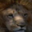}\includegraphics[width=0.4cm]{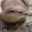}\includegraphics[width=0.4cm]{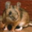}\includegraphics[width=0.4cm]{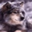}\includegraphics[width=0.4cm]{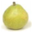}\includegraphics[width=0.4cm]{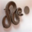}\includegraphics[width=0.4cm]{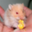}\includegraphics[width=0.4cm]{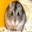}\includegraphics[width=0.4cm]{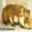}\includegraphics[width=0.4cm]{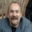}\includegraphics[width=0.4cm]{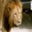}\includegraphics[width=0.4cm]{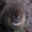}\includegraphics[width=0.4cm]{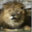}\includegraphics[width=0.4cm]{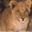}\\
		\hline   18: streetcar,train,bridge,bus  & \includegraphics[width=0.4cm]{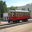}\includegraphics[width=0.4cm]{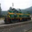}\includegraphics[width=0.4cm]{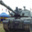}\includegraphics[width=0.4cm]{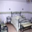}\includegraphics[width=0.4cm]{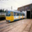}\includegraphics[width=0.4cm]{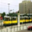}\includegraphics[width=0.4cm]{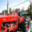}\includegraphics[width=0.4cm]{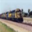}\includegraphics[width=0.4cm]{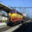}\includegraphics[width=0.4cm]{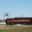}\includegraphics[width=0.4cm]{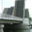}\includegraphics[width=0.4cm]{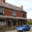}\includegraphics[width=0.4cm]{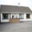}\includegraphics[width=0.4cm]{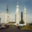}\includegraphics[width=0.4cm]{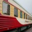}\includegraphics[width=0.4cm]{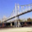}\includegraphics[width=0.4cm]{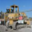}\includegraphics[width=0.4cm]{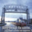}\includegraphics[width=0.4cm]{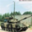}\includegraphics[width=0.4cm]{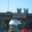}\\
		\hline   19: oak,maple,poppy,sunflower  & \includegraphics[width=0.4cm]{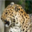}\includegraphics[width=0.4cm]{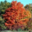}\includegraphics[width=0.4cm]{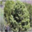}\includegraphics[width=0.4cm]{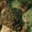}\includegraphics[width=0.4cm]{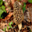}\includegraphics[width=0.4cm]{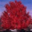}\includegraphics[width=0.4cm]{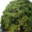}\includegraphics[width=0.4cm]{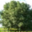}\includegraphics[width=0.4cm]{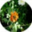}\includegraphics[width=0.4cm]{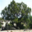}\includegraphics[width=0.4cm]{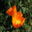}\includegraphics[width=0.4cm]{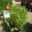}\includegraphics[width=0.4cm]{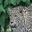}\includegraphics[width=0.4cm]{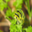}\includegraphics[width=0.4cm]{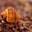}\includegraphics[width=0.4cm]{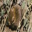}\includegraphics[width=0.4cm]{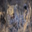}\includegraphics[width=0.4cm]{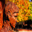}\includegraphics[width=0.4cm]{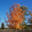}\includegraphics[width=0.4cm]{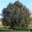}\\
		\hline
	\end{tabular}
\end{table*}

\section{Conclusions}
\label{sec:conclusions}
In this paper, we introduced an unsupervised framework for exploration of the representations learned by a CNN, based on constrained and regularized coupled matrix factorization. Our proposed method is unique and novel in that it is the first such framework to allow for {\em joint} exploration of the representations that a CNN has learned across features (pixels), activations, and data instances.  This is in stark contrast to existing state-of-the-art works, which are typically restricted to one of those three modalities, as summarized in Fig. \ref{fig:comp}. Furthermore, owing to the simplicity of the factorization model, our method can provide easily interpretable insights. As a result, our proposed framework offers maximum flexibility and bridges the gap between existing works, while  producing comparable results to state-of-the-art, when used for the same (albeit limited) purpose of existing work. Case in point, in this paper, we demonstrate a number of applications of our framework drawing parallels to what existing work can offer compared to our results, including the extraction of instance-based interpretable concepts (Sec. \ref{sec:concepts}), and based on those concepts we provide insights on the the behavior of neurons in different layers (Sec. \ref{sec:activations}), and instance-level pixel-based insights (Sec. \ref{sec:pixels}). In future work, we will investigate the adaptation to our framework to different architectures (e.g., RNN and GCN) and different applications (e.g., NLP, Graph Mining, and Recommendation Systems).

\hide{
\section{Acknowledgements}
{\scriptsize
Research was supported by the National Science Foundation Grant No. XXXXXX. Any opinions, findings, and conclusions or recommendations expressed in this material are those of the author(s) and do not necessarily reflect the views of the funding parties.
}
}
\balance
\bibliographystyle{plain}
\bibliography{BIB/vagelis_refs.bib}

\end{document}